\documentclass{Interspeech2024}

\interspeechcameraready 
\usepackage{tipa}
\usepackage{tabularx, booktabs, makecell, caption}
\usepackage{siunitx}

\title{A layer-wise analysis of Mandarin and English suprasegmentals in SSL speech models}

\name{Antón}{de la Fuente}
\name{Dan}{Jurafsky}

\address{Stanford University}
\email{antondlf@stanford.edu, jurafsky@stanford.com}


\begin{document}

\maketitle


\begin{abstract}
    

    This study asks how self-supervised speech models represent suprasegmental categories like Mandarin lexical tone, English lexical stress, and English phrasal accents. Through a series of probing tasks, we make layer-wise comparisons of English and Mandarin 12 layer monolingual models. Our findings suggest that 1) English and Mandarin wav2vec 2.0 models learn contextual representations of abstract suprasegmental categories which are strongest in the middle third of the network. 2) Models are better at representing features that exist in the language of their training data, and this difference is driven by enriched context in transformer blocks, not local acoustic representation. 3) Fine-tuned wav2vec 2.0 improves performance in later layers compared to pre-trained models mainly for lexically contrastive features like tone and stress,  4) HuBERT and WavLM learn similar representations to wav2vec 2.0, differing mainly in later layer performance. Our results extend previous understanding of how models represent suprasegmentals and offer new insights into the language-specificity and contextual nature of these representations.

\end{abstract}

\section{Introduction}

  How do self-supervised learning (SSL) speech models represent suprasegmental information about features like accents, stresses, and lexical tones? Probing how these models build up these linguistic representations over layers could help us understand how they incorporate contextual information, how  their representations differ across languages, and how these representations get modified when fine-tuned.
  
  Here we study three models: wav2vec 2.0 \cite{Baevski2020wav2vec2A}, HuBERT \cite{Hsu2021HuBERTSS}, and WavLM \cite{WavLM}, to understand how they represent  Mandarin lexical tone, English lexical stress, and English phrasal pitch accents.  Comparing stress and tone, which are properties of words, with  pitch accent, which is phrasal, allows us to explore the role of the lexicon in representation learning.   

 We study these models through probing: training classifiers to predict linguistic classes from internal representations of the model to see what kinds of information the model represents about those classes. Probing has been used to show that SSL models represent phone identity  \cite{Martin2023ProbingSS, CormacEnglish2022DomainInformedPO, Ma2020ProbingAR, Pasad2023WhatDS, Pasad2022ComparativeLA, Pasad2021LayerWiseAO}, and semantic or syntactic features of words \cite{Pasad2021LayerWiseAO, Pasad2023WhatDS, Shah2021WhatAD}. A consistent finding is that wav2vec 2.0 follows an autoencoder-like behavior where early and later layers relate more closely to local acoustic features and the middle layers seem to represent more abstract linguistic categories. HuBERT and WavLM, on the other hand, have representations that relate closely to abstract linguistic categories up to their last layers.
 \cite{Martin2023ProbingSS, Pasad2023WhatDS, Pasad2022ComparativeLA}.
 While we also know that suprasegmentals like prosody and tone are represented by SSL models \cite{Yuan2021AutomaticRO, Yang2023WhatCA, prosodyTasks, emotionProbing},  we know less about how these representations develop across layers, what contexts they draw on, and how language-general they are. Our investigations of how suprasegmental representations develop through model layers led us to the following contributions:
 
    \begin{enumerate}
        \item wav2vec 2.0 learns representations of English stress, Mandarin tone, and English phrasal accents. The best classifier probe performance uses representations that come from the middle third of the network. Layer-wise improvements in probe performance are not driven by improvements in the ability of the model to better track F0. This result suggests that categorical suprasegmental representations are more abstract and not directly tied to simple acoustic features.
        \item  Representation performance is similar for all models at the CNN output. As context gets added by the transformer, models improve at classifying features of the language they were trained on. This shows that language specificity in SSL models is driven by domain specific context and not accurate acoustic representation.
        \item Fine-tuning wav2vec2.0 for ASR improves performance in later layers even if probed features are not orthographically represented. This appears more robust for lexical categories (i.e., tone and stress) than for phrasal ones like accent. Probe performance here likely improves because of the model's improved knowledge of lexical identity, which explains why the effect is larger for lexically contrastive suprasegmentals like stress and tone.
        \item HuBERT, WavLM, and wav2vec 2.0 represent suprasegmental categories equally well and show similar layer-wise behavior.
        
    \end{enumerate}

\section{Models \& corpora}

\subsection{Models probed}

    The main analysis was performed on two models: wav2vec2-base \cite{Baevski2020wav2vec2A}, and mandarin-wav2vec2 \cite{Lu2022ACK}. The models are trained on comparable datasets ($\sim{960}$h of read speech), and have somewhat comparable ASR fine-tuned versions (trained on 100h and 175h respectively). English HuBert \cite{Hsu2021HuBERTSS} and WavLM \cite{WavLM} versions trained on the same data as wav2vec2-base were also probed to assess whether the pre-training task affect representations. Models details summarized in table  1.

    All models consist of a feature extraction network made up of 7 CNN layers and then a context network consisting of 12 transformer blocks. These are trained through a masked prediction task with masked tokens derived from quantizing CNN output (wav2vec 2.0), representations of earlier iterations of the model (HuBERT), or noisy/overlapping audio (WavLM).

    \begin{table}[t]
      \caption{Model Summary}
      \label{tab:table2}
      \centering
      \begin{tabular}{ l l l}
        \toprule
        \textbf{Model Name} & \textbf{Train Set Size} & \textbf{Language}\\
        \midrule
        wav2vec2-base              &  $\sim960$h & English ~~~             \\
        wav2vec2-base-100h         &  $\sim960$h $+ 100$h & English~~~             \\
        mandarin-wav2vec2          &  $\sim960$h  & Mandarin~~~             \\
        mandarin-w2v2-aishell1 &  $\sim960$h $+175$h & Mandarin~~~             \\
        HuBERT                     & $\sim960$h  & English ~~~             \\
        WavLM                      &  $\sim960$h  & English ~~~             \\
        
        \bottomrule
      \end{tabular}
  
\end{table}

\subsection{Corpora}
    For English tasks, we used 75 conversations from the Switchboard corpus \cite{switchboard} that had been annnotated with stress and ToBI accent labels from NXT \cite{nxtswitchboard}. This yielded a total of 7.5 hours of audio with a mean clip duration of 5.5 seconds.
    
    For Mandarin tasks we used Global Timit Mandarin-Chinese (GTMC), which consists of 50 speakers reading 120 sentences each. Word alignment annotations in Chinese characters and onset-rhyme alignments were provided \cite{globaltimitmandarin}. Pinyin transcriptions of Chinese characters were automatically derived and then tones from those transcriptions were mapped to each syllable. This corpus has a total of 5.7 hours of audio with a mean clip duration of 3.4 seconds.
    
\section{Probing tasks}

\subsection{Probing methodology}

    For all models and tasks, we fit an independent probe on the output of the convolutional feature extractor (layer 0) and each transformer layer (layers 1-12)\footnote{Code available at \url{https://github.com/antondlf/prosody_probing}}. Our analyses were conducted using linear probes following \cite{Ma2020ProbingAR}\footnote{Also following \cite{Ma2020ProbingAR}, we trained MLP probes. Although they perform slightly better overall they have similar layer-wise behavior so we omit them for brevity.}. For the regression task, we fit linear least squares regressions. For classifications tasks, we fit either binary or multinomial logistic regressions with L1 regularization, a threshold of 0.5, a C parameter of 1, and the saga solver \cite{Defazio2014SAGAAF}. Instead of tuning hyperparameters they were kept constant throughout all layers and tasks, and with the exception of the regularization penalty type no changes were made throughout the development process. L1 regularization was selected after an initial failed run because logistic regressions with L2 regularization could not converge. 
    
    Probes were fit frame-by-frame using the entire 768 dimensional embeddings as input features, with no pooling of representations. The train test split was performed by speaker, with $80\%$ of speakers' audio used for training and the remaining $20\%$ of speakers held out for evaluation. Classifier probes were evaluated using F1 scores (macro-averaged for the non-binary case). Regression probes were evaluated using R-squared\footnote{Results show F0 R-squared evals and classification F1 scores for both binary and multi-class tasks. \textbf{Note that these metrics are not meant to be compared across tasks}. Only relative layer-wise differences are meaningful and interpretable.}.

\subsection{Task setup}
    \begin{table}[t]

      \caption{Task label distribution (labeled syllables)}
      \label{tab:table3}
      \centering
      \begin{tabular}{ l l l}
        \toprule
        \textbf{Label} & \textbf{Train} & \textbf{Test} \\
        \midrule
        \textit{English Stress  (total syllables)} & $89,452$ & $21,965$ \\
        \midrule
        Stress   & $70.4\%$ & $69.5\%$ ~~~   \\
        No Stress (Positive class)     & $29.6\%$ & $30.5\%$ ~~~   \\
        \midrule
        \textit{English Accents (total syllables)} & $89,453$ & $21,965$ \\
        \midrule
        Accent (Positive class)     & $29.1\%$ & $28.3\%$ ~~~   \\
        No Accent   & $70.9\%$ & $71.7\%$ ~~~   \\
        \midrule
        \textit{Mandarin Tone (total syllables)} & $132,467$ & $33,158$\\
        \midrule
        Tone 1      & $22.2\%$ & $22.3\%$ ~~~   \\
        Tone 2      & $21.2\%$ & $21.2\%$ ~~~   \\
        Tone 3      & $17.4\%$ & $17.1\%$ ~~~   \\
        Tone 4      & $33.3\%$ & $33.4\%$ ~~~   \\
        Tone 5 (neutral)& $5.9\%$ & $5.9\%$ ~~~   \\
        \bottomrule
      \end{tabular}

\end{table}

    Each clip from our data was passed through each of our networks (from table 1). The embeddings of layers 0-12 were saved to use as input features to the probes. Time-stamps from provided corpus annotations were used to align each label with its corresponding model embedding. For the classification tasks, all frames falling within a given syllable had the same label. All non-word frames were excluded from the data. Task specific information is given below, with label distribution summarized in table 2\footnote{Reported proportions are on syllable counts, actual data point counts are about $10$ times greater because each syllable contains an average of $10$ frames.}.
    
    \paragraph*{\textbf{English stress}} Stress labels from NXT Switchboard distinguish between primary stress 'p', secondary stress 's', and unstressed 'n', but we collapsed secondary stress and unstressed into a single unstressed category to create a binary task. We did not exclude monosyllabic words from the data because there is a distinction in the labels between stressed and unstressed monosyllables. This led the unstressed category to be the least common. Reported F1 scores for the lexical stress task therefore treat unstressed syllables as the positive class. 
    
    \paragraph*{\textbf{English pitch accents}} NXT Switchboard contains pitch accents labeled at individual time points. We spread these labels to the syllable domain, labeling all frames in a syllable containing an accent as accented. Accent distinctions were collapsed to create a binary (accented/unaccented) task. The minority class was accented syllables, which was used as the positive class for F1 evaluation.
    
    \paragraph*{\textbf{Mandarin tone}} Mandarin tone annotations were automatically derived from Chinese word level character annotations provided for Global-Timit Mandarin Chinese (GTMC) using a python package for pinyin transcription. Tone indices were then mapped to each syllable using alignments provided with the corpus. This created a 5 way classification task, with the 4 Mandarin tones and neutral tone. 
    \paragraph*{\textbf{F0}} For both corpora, F0 was extracted using the autocorrelation function from Praat's python API, Parselmouth \cite{Boersma2003PraatDP, parselmouth}. Each frame was then assigned a pitch measure in Hertz\footnote{Mel transformed F0 performed similarly and will not be reported.}. Though both GTMC and Switchboard were used for this task, only the Switchboard results will be shown for brevity. GTMC probes performed marginally better than Switchboard ones, but layer-wise progression is similar.

\section{Results \& discussion}
    \begin{figure}[t]
            \centering
            \includegraphics[width=0.44\textwidth]{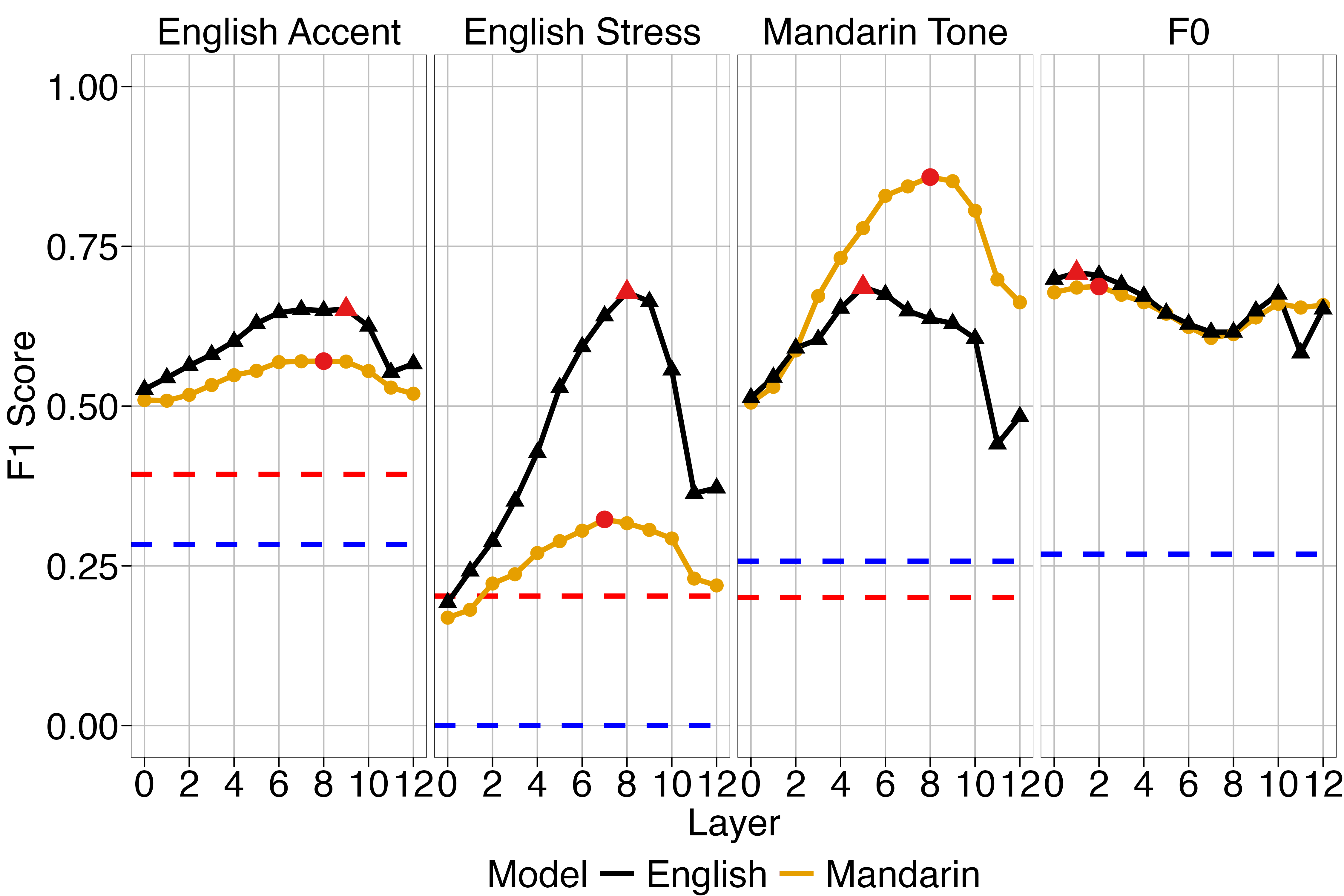}
            \caption{\protect\raggedright English monolingual wav2vec2-base (black) and Mandarin monolingual mandarin-wav2vec2 (orange) probe performance on each task. Red point indicates the best layer for each model. Dashed lines are random baselines (red), or Mel-Filterbank baselines (blue)}. F0 task score is R-Squared, all other tasks report F1 scores.
            \label{fig:figure1}
    \end{figure}
    
    Language comparisons are shown in Figures 1 and 2. When discussing these results wav2vec2-base will be referred to as the English model and mandarin-wav2vec2 as the Mandarin model. Their ASR fine-tuned counterparts will just be referred to as the English or Mandarin fine-tuned model. Figure 3 shows a comparisons between wav2vec 2.0, HuBERT, and WavLM.

    \subsection*{\textbf{English and Mandarin wav2vec 2.0 comparison}}
    \paragraph*{\textbf{Result 1.1}} \textit{wav2vec 2.0 representations of abstract suprasegmental categories are better in the middle layers than in early and late layers.}  Model representations matched to the test language reach their best performance in layers 8 or 9 (Figure 1). That is, the English model (black line) for the English stress and accent tasks peaks in layers 8 and 9 respectively. The Mandarin model (orange line) for the Mandarin tone task peaks in layer 8. The fact that models trained on the other language also peak in the middle layers suggests that the representations of suprasegmental categories develop in analogous ways across languages.

    \paragraph*{\textbf{Result 1.2}} \textit{Model representations of suprasegmental categories appear to be independent of their ability to track F0 accurately.} Performance for the F0 task (panel 4) fluctuates through the context network and performs similarly for both the English and Mandarin models. Both models reach their lowest performance in layer 7\footnote{This is not strictly true for the English model, which drops significantly for layer 11 and recovers in layer 12. This may be an anomaly of this specific model and not representative of any trend in the layer-wise development of representations.}, a layer earlier than layers where the representations peak for the classification tasks. This indicates that performance improvements are not driven by improved representation of the acoustic signal and implies that model representations are sensitive to abstract linguistic structure and not just acoustic features.

    \paragraph*{\textbf{Result 2}} \textit{Models develop language specific representations of suprasegmental categories only in the context network.} Figure 1 shows that representations of suprasegmentals improve in through the context network. This improvement is larger when the model's pre-training language is matched to the task language. That is, English model representations (black lines) improve more and more quickly than Mandarin model representations (orange line) for both English accent and stress (panels 1 and 2 respectively). The opposite is true for the Mandarin tone task (panel 3). Crucially, layer 0 performance is similar for all models, implying that language specific information is encoded only in the context network.

    \begin{figure}
        \centering
        \includegraphics[width=0.44\textwidth]{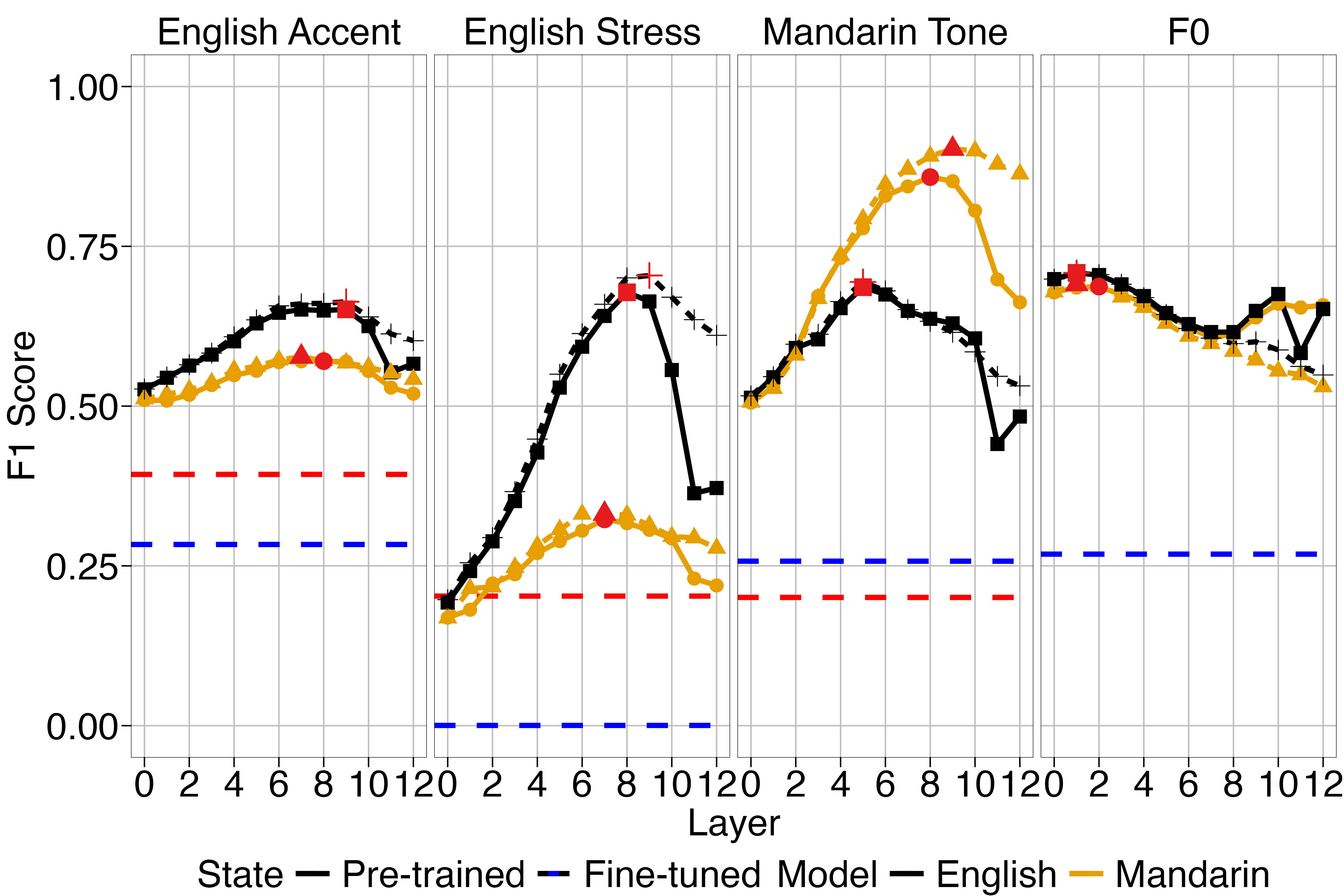}
        \caption{Fine-tuned (dashed) and pre-trained (solid) monolingual English (black) and Mandarin (orange) wav2vec 2.0 model performance on all tasks. Red point indicates the best layer for each model. Dashed lines are random baselines (red), or Mel-Filterbank baselines (blue)}.
        \label{fig:figure2}
    \end{figure}
    \subsection*{\textbf{ASR fine-tuning effects}} 
    \paragraph*{\textbf{Result 3.1}}\textit{Fine-tuning for ASR improves suprasegmental representation, especially for models matched to the task language.} Figure 2 shows performance of fine-tuned models (dashed lines) relative to their pre-trained counter parts. Fine tuning improves performance for all models matched to the task language, especially in the last four layers. This is consistent with findings in \cite{Pasad2021LayerWiseAO}. It is not clear that the effect holds for models that are not matched to the task language.

    \paragraph*{\textbf{Result 3.2}}\textit{The fine-tuning performance boost is stronger for suprasegmentals that are distinctive at the word level.} The English model improves for stress after fine-tuning, and its peak layer moves to layer 9. The Mandarin model improves for tone after fine-tuning and its peak layer moves to layer 10. This is likely due to the fact that stress and tone information is implicitly encoded by orthography, so the fine-tuned ASR model has learned about their distributions indirectly. The phrasal accent task shows some improvements, but the effect is less strong and the peak layer does not change. This may indicate that this feature is not implicitly encoded by the orthography task.
    \begin{figure}
        \centering
        \includegraphics[width=0.44\textwidth]{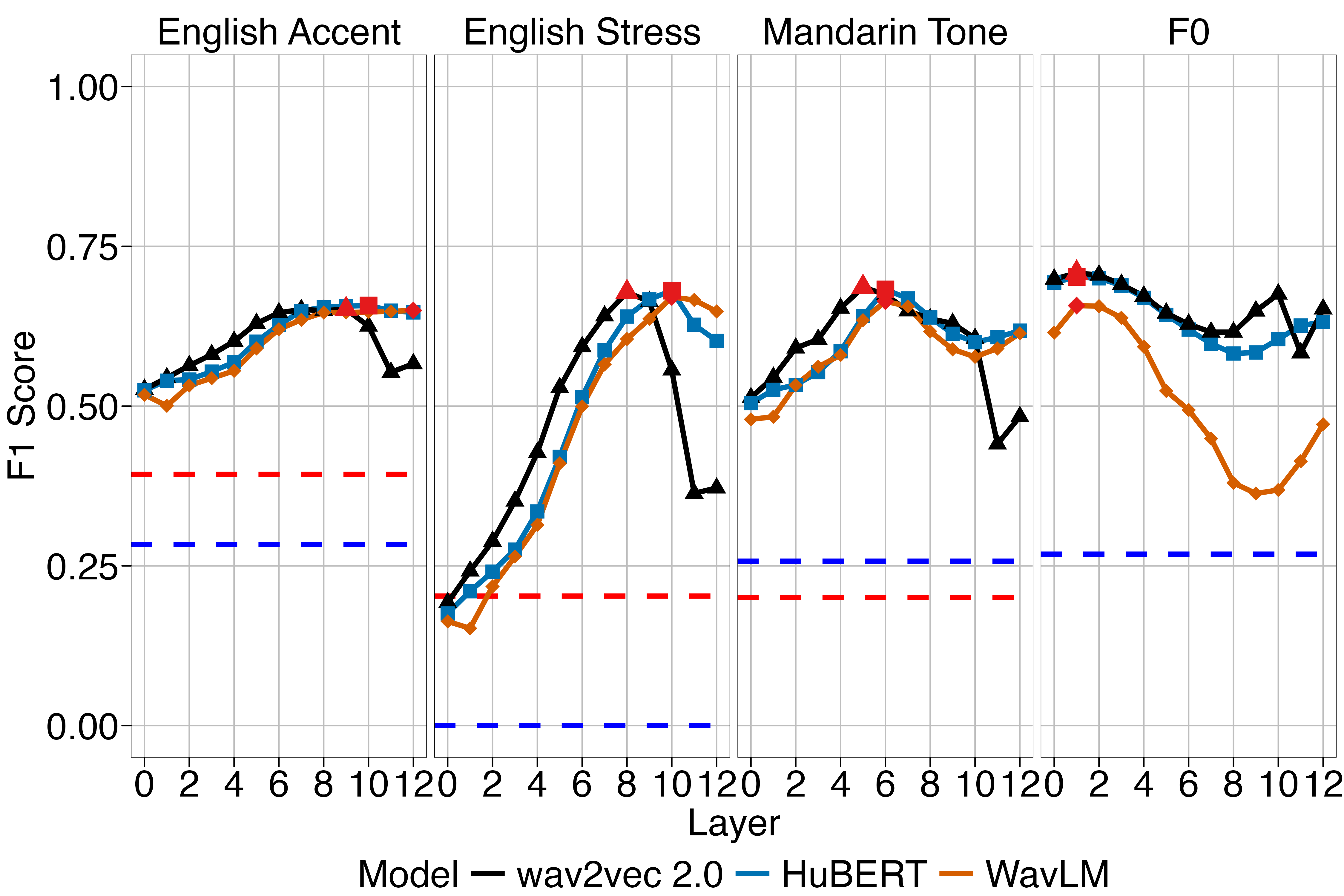}
        \caption{English wav2vec 2.0 (black) HuBERT (tawny) and WavLM (blue) model performance on all tasks. The best layer for each model has a red point. Dashed lines are random baselines (red), or Mel-Filterbank baselines (blue)}.
        \label{fig:figure3}
    \end{figure}

    \subsection*{\textbf{HuBERT and WavLM results}} 
    \paragraph*{\textbf{Result 4.1}} \textit{Pre-text task differences} do not affect representations of suprasegmentals outside expected layer-wise differences. Figure 2 shows that HuBERT and WavLM perform similarly to wav2vec 2.0 on all probing tasks. The main difference is that the last few layers do not drop in performance as drastically, which is expected given \cite{Pasad2022ComparativeLA}. This indicates that our results for wav2vec 2.0 are likely due to the architecture and independent of the kind of masked modeling pre-text task used. The lack of pre-trained monolingual models that are not in English prevents us from testing this hypothesis directly, so we used this experiment as the closest feasible proxy.
    
    \paragraph*{\textbf{Result 4.2}} \textit{HuBERT is the best performing model for the English accent and English stress tasks.} For Mandarin tone, the English wav2vec 2.0 model performs best. Though further work is needed, this suggests that the wav2vec 2.0 pre-training task might be more suitable for language transferable suprasegmental representations. WavLM is the only model to perform worse in layer 1 than layer 0, though only for the English stress and English accent tasks.

    \subsection{\textbf{General discussion}}



    Result 4 suggests that findings for wav2vec 2.0 are similar to those for HuBERT and WavLM. Though this study only shows this for the English models, the similarities among the model performances and layer-wise behavior suggest that they might behave similarly if pre-trained on Mandarin as well. Thus, for the purposes of this discussion we will treat our results as applying to SSL models in general, even if that scope is somewhat speculative.

    Result 1 suggests that the representations leveraged by probes to classify suprasegmental task labels are abstract. That is, they do not have a direct linear relationship to surface acoustic features despite the fact that F0 is a significant cue for all three categories \cite{lehiste1970suprasegmentals, accentCorrelates, tupper_characterizing_2020}.  Result 2 shows that changing the pre-training language does little to layer 0 performance but has a large effect on all other layers\footnote{Though not shown here, multilingual models exhibit similar behavior.}. This suggests the context network as the site for language specific learning.

    Despite the fact that Mandarin and English have radically different orthographies, the effects of ASR fine-tuning on tone and stress respectively are notable. Neither feature is encoded orthographically, yet training for orthographic transcription clearly boosts model representations of these features. This is likely because fine-tuning for ASR emphasizes lexical information, and both of these features are lexical. This effect is less clear for English accents, indicating that their representation at the lexical level is weaker.

    In general, the English phrasal accent task behaves differently when compared to stress and tone. This is indicative of a divergence between phrasal and lexical representations. Because the phrasal context is wider and these probes were fit on individual frames, it is possible that this divergence is due to an architectural limitation on the width of the context that is relevant to a given token's suprasegmental status. Further work is necessary to verify this claim.
   
    Lastly, the tone task is worth mentioning on its own since it has the largest difference in what layers achieve the best performance between the English and Mandarin models. The English model is fairly good at classifying tone, consistent with \cite{Yuan2021AutomaticRO}. It also has its best performing layer earlier in the network than the Mandarin model and drops off steeply. Earlier layers tend to achieve high performance for phone classification, which suggests that this model's representation of the phonemic domain is more suitable to capture the structure of Mandarin tone. This is consistent with findings in \cite{CormacEnglish2022DomainInformedPO, Pasad2021LayerWiseAO}, and the fine-tuning results in \cite{Yuan2021AutomaticRO}, where data augmentation through English phone recognition yielded better results than fine-tuning on Mandarin tone recognition alone.
    \section{Limitations and further work}

    L1-penalty reduced convergence issues for these tasks, but did not entirely eliminate them. Specifically, this continued to be an issue for a few layers of the fine-tuned models on the tone task. Our data also presented some limitations in comparability, since switchboard data is upsampled conversational speech and GTMC is read speech. This created confounds that we tried to avoid by reducing comparisons across tasks in our interpretations. 

    Conditioning out earlier layer representations following \cite{hewitt-etal-2021-conditional} would also improve the reliability of results. This would control for the possibility of models learning non-overlapping information about each category through the layers. Further improvements to the clarity of the results might be achieved by conditioning out or controlling for lexical identity in the experiment setup as well.

\section{Conclusion}

    This study showed that SSL models of speech learn representations of abstract suprasegmental categores like Mandarin tone, English stress, and English phrasal pitch accents. These representations are not directly related to local acoustic information, and reach their best performance in deep network layers. Performance differences based on pre-training language arise only in the context network, implying that all models develop similar local representations of the signal. Fine-tuning for ASR improves representations especially for lexical categories like tone and stress. Lastly, tests of models with three different masked modeling pre-training tasks performed similarly, implying that the findings hold more generally for this class of SSL speech models.

\paragraph*{\textbf{Acknowledgments}:} Thank you to Karen Livescu, Martijn Bartelds, and Nay San for support throughout the development and writing process. Thank you also to Arto Anttila, Chris Manning, and the Jurafsky Lab.

\bibliographystyle{IEEEtran}
\bibliography{mybib}

\begin{thebibliography}{10}
\providecommand{\url}[1]{#1}
\csname url@samestyle\endcsname
\providecommand{\newblock}{\relax}
\providecommand{\bibinfo}[2]{#2}
\providecommand{\BIBentrySTDinterwordspacing}{\spaceskip=0pt\relax}
\providecommand{\BIBentryALTinterwordstretchfactor}{4}
\providecommand{\BIBentryALTinterwordspacing}{\spaceskip=\fontdimen2\font plus
\BIBentryALTinterwordstretchfactor\fontdimen3\font minus \fontdimen4\font\relax}
\providecommand{\BIBforeignlanguage}[2]{{%
\expandafter\ifx\csname l@#1\endcsname\relax
\typeout{** WARNING: IEEEtran.bst: No hyphenation pattern has been}%
\typeout{** loaded for the language `#1'. Using the pattern for}%
\typeout{** the default language instead.}%
\else
\language=\csname l@#1\endcsname
\fi
#2}}
\providecommand{\BIBdecl}{\relax}
\BIBdecl

\bibitem{Baevski2020wav2vec2A}
\BIBentryALTinterwordspacing
A.~Baevski, H.~Zhou, A.~rahman Mohamed, and M.~Auli, ``wav2vec 2.0: A framework for self-supervised learning of speech representations,'' \emph{ArXiv}, vol. abs/2006.11477, 2020. [Online]. Available: \url{https://api.semanticscholar.org/CorpusID:219966759}
\BIBentrySTDinterwordspacing

\bibitem{Hsu2021HuBERTSS}
\BIBentryALTinterwordspacing
W.-N. Hsu, B.~Bolte, Y.-H.~H. Tsai, K.~Lakhotia, R.~Salakhutdinov, and A.~rahman Mohamed, ``Hubert: Self-supervised speech representation learning by masked prediction of hidden units,'' \emph{IEEE/ACM Transactions on Audio, Speech, and Language Processing}, vol.~29, pp. 3451--3460, 2021. [Online]. Available: \url{https://api.semanticscholar.org/CorpusID:235421619}
\BIBentrySTDinterwordspacing

\bibitem{WavLM}
\BIBentryALTinterwordspacing
S.~Chen, C.~Wang, Z.~Chen, Y.~Wu, S.~Liu, Z.~Chen, J.~Li, N.~Kanda, T.~Yoshioka, X.~Xiao, J.~Wu, L.~Zhou, S.~Ren, Y.~Qian, Y.~Qian, M.~Zeng, and F.~Wei, ``Wavlm: Large-scale self-supervised pre-training for full stack speech processing,'' \emph{IEEE Journal of Selected Topics in Signal Processing}, vol.~16, pp. 1505--1518, 2021. [Online]. Available: \url{https://api.semanticscholar.org/CorpusID:239885872}
\BIBentrySTDinterwordspacing

\bibitem{Martin2023ProbingSS}
\BIBentryALTinterwordspacing
K.~Martin, J.~Gauthier, C.~Breiss, and R.~P. Levy, ``Probing self-supervised speech models for phonetic and phonemic information: a case study in aspiration,'' \emph{ArXiv}, vol. abs/2306.06232, 2023. [Online]. Available: \url{https://api.semanticscholar.org/CorpusID:259137399}
\BIBentrySTDinterwordspacing

\bibitem{CormacEnglish2022DomainInformedPO}
\BIBentryALTinterwordspacing
P.~C. English, J.~D. Kelleher, and J.~Carson-Berndsen, ``Domain-informed probing of wav2vec 2.0 embeddings for phonetic features,'' \emph{Proceedings of the 19th SIGMORPHON Workshop on Computational Research in Phonetics, Phonology, and Morphology}, 2022. [Online]. Available: \url{https://api.semanticscholar.org/CorpusID:250390755}
\BIBentrySTDinterwordspacing

\bibitem{Ma2020ProbingAR}
\BIBentryALTinterwordspacing
D.~Ma, N.~Ryant, and M.~Y. Liberman, ``Probing acoustic representations for phonetic properties,'' \emph{ICASSP 2021 - 2021 IEEE International Conference on Acoustics, Speech and Signal Processing (ICASSP)}, pp. 311--315, 2020. [Online]. Available: \url{https://api.semanticscholar.org/CorpusID:225067148}
\BIBentrySTDinterwordspacing

\bibitem{Pasad2023WhatDS}
\BIBentryALTinterwordspacing
A.~Pasad, C.-M. Chien, S.~Settle, and K.~Livescu, ``What do self-supervised speech models know about words?'' \emph{ArXiv}, vol. abs/2307.00162, 2023. [Online]. Available: \url{https://api.semanticscholar.org/CorpusID:259316239}
\BIBentrySTDinterwordspacing

\bibitem{Pasad2022ComparativeLA}
\BIBentryALTinterwordspacing
A.~Pasad, B.~Shi, and K.~Livescu, ``Comparative layer-wise analysis of self-supervised speech models,'' \emph{ICASSP 2023 - 2023 IEEE International Conference on Acoustics, Speech and Signal Processing (ICASSP)}, pp. 1--5, 2022. [Online]. Available: \url{https://api.semanticscholar.org/CorpusID:253397618}
\BIBentrySTDinterwordspacing

\bibitem{Pasad2021LayerWiseAO}
\BIBentryALTinterwordspacing
A.~Pasad, J.-C. Chou, and K.~Livescu, ``Layer-wise analysis of a self-supervised speech representation model,'' \emph{2021 IEEE Automatic Speech Recognition and Understanding Workshop (ASRU)}, pp. 914--921, 2021. [Online]. Available: \url{https://api.semanticscholar.org/CorpusID:235795805}
\BIBentrySTDinterwordspacing

\bibitem{Shah2021WhatAD}
\BIBentryALTinterwordspacing
J.~Shah, Y.~K. Singla, C.~Chen, and R.~R. Shah, ``What all do audio transformer models hear? probing acoustic representations for language delivery and its structure,'' \emph{ArXiv}, vol. abs/2101.00387, 2021. [Online]. Available: \url{https://api.semanticscholar.org/CorpusID:230433693}
\BIBentrySTDinterwordspacing

\bibitem{Yuan2021AutomaticRO}
\BIBentryALTinterwordspacing
J.~Yuan, N.~Ryant, X.~Cai, K.~W. Church, and M.~Y. Liberman, ``Automatic recognition of suprasegmentals in speech,'' \emph{ArXiv}, vol. abs/2108.01122, 2021. [Online]. Available: \url{https://api.semanticscholar.org/CorpusID:236881600}
\BIBentrySTDinterwordspacing

\bibitem{Yang2023WhatCA}
\BIBentryALTinterwordspacing
M.~Yang, R.~C.~C. Shekar, O.~Kang, and J.~H.~L. Hansen, ``What can an accent identifier learn? probing phonetic and prosodic information in a wav2vec2-based accent identification model,'' \emph{ArXiv}, vol. abs/2306.06524, 2023. [Online]. Available: \url{https://api.semanticscholar.org/CorpusID:259137711}
\BIBentrySTDinterwordspacing

\bibitem{prosodyTasks}
G.-T. Lin, C.-L. Feng, W.-P. Huang, Y.~Tseng, T.-H. Lin, C.-A. Li, H.-y. Lee, and N.~G. Ward, ``On the utility of self-supervised models for prosody-related tasks,'' in \emph{2022 IEEE Spoken Language Technology Workshop (SLT)}, 2023, pp. 1104--1111.

\bibitem{emotionProbing}
Y.~Li, Y.~Mohamied, P.~Bell, and C.~Lai, ``Exploration of a self-supervised speech model: A study on emotional corpora,'' in \emph{2022 IEEE Spoken Language Technology Workshop (SLT)}, 2023, pp. 868--875.

\bibitem{Lu2022ACK}
\BIBentryALTinterwordspacing
K.~Lu and K.-Y. Chen, ``A context-aware knowledge transferring strategy for ctc-based asr,'' \emph{2022 IEEE Spoken Language Technology Workshop (SLT)}, pp. 60--67, 2022. [Online]. Available: \url{https://api.semanticscholar.org/CorpusID:252846706}
\BIBentrySTDinterwordspacing

\bibitem{switchboard}
J.~Godfrey, E.~Holliman, and J.~McDaniel, ``Switchboard: telephone speech corpus for research and development,'' in \emph{[Proceedings] ICASSP-92: 1992 IEEE International Conference on Acoustics, Speech, and Signal Processing}, vol.~1, 1992, pp. 517--520 vol.1.

\bibitem{nxtswitchboard}
\BIBentryALTinterwordspacing
S.~Calhoun, J.~Carletta, J.~M. Brenier, N.~Mayo, D.~Jurafsky, M.~Steedman, and D.~Beaver, ``The nxt-format switchboard corpus: a rich resource for investigating the syntax, semantics, pragmatics and prosody of dialogue,'' \emph{Language Resources and Evaluation}, vol.~44, no.~4, pp. 387--419, 2010. [Online]. Available: \url{http://www.jstor.org/stable/40925580}
\BIBentrySTDinterwordspacing

\bibitem{globaltimitmandarin}
\BIBentryALTinterwordspacing
H.~Ding, S.~Liao, Y.~Zhan, J.~Yuan, and M.~Liberman, ``{Global TIMIT Mandarin Chinese},'' 2021. [Online]. Available: \url{https://hdl.handle.net/11272.1/AB2/2CCXH8}
\BIBentrySTDinterwordspacing

\bibitem{Defazio2014SAGAAF}
\BIBentryALTinterwordspacing
A.~Defazio, F.~R. Bach, and S.~Lacoste-Julien, ``Saga: A fast incremental gradient method with support for non-strongly convex composite objectives,'' in \emph{Neural Information Processing Systems}, 2014. [Online]. Available: \url{https://api.semanticscholar.org/CorpusID:218654665}
\BIBentrySTDinterwordspacing

\bibitem{Boersma2003PraatDP}
\BIBentryALTinterwordspacing
P.~Boersma and D.~Weenink, ``Praat: doing phonetics by computer,'' 2003. [Online]. Available: \url{https://api.semanticscholar.org/CorpusID:60594797}
\BIBentrySTDinterwordspacing

\bibitem{parselmouth}
Y.~Jadoul, B.~Thompson, and B.~de~Boer, ``Introducing {P}arselmouth: A {P}ython interface to {P}raat,'' \emph{Journal of Phonetics}, vol.~71, pp. 1--15, 2018.

\bibitem{lehiste1970suprasegmentals}
I.~Lehiste, \emph{Suprasegmentals.}\hskip 1em plus 0.5em minus 0.4em\relax Massachusetts Inst. of Technology P, 1970.

\bibitem{accentCorrelates}
Q.~Yan, S.~Vaseghi, D.~Rentzos, C.-H. Ho, and E.~Turajlic, ``Analysis of acoustic correlates of british, australian and american accents,'' in \emph{2003 IEEE Workshop on Automatic Speech Recognition and Understanding (IEEE Cat. No.03EX721)}, 2003, pp. 345--350.

\bibitem{tupper_characterizing_2020}
\BIBentryALTinterwordspacing
P.~Tupper, K.~Leung, Y.~Wang, A.~Jongman, and J.~A. Sereno, ``Characterizing the distinctive acoustic cues of {Mandarin} tones,'' \emph{The Journal of the Acoustical Society of America}, vol. 147, no.~4, pp. 2570--2580, Apr. 2020, \_eprint: https://pubs.aip.org/asa/jasa/article-pdf/147/4/2570/14124676/2570\_1\_online.pdf. [Online]. Available: \url{https://doi.org/10.1121/10.0001024}
\BIBentrySTDinterwordspacing

\bibitem{hewitt-etal-2021-conditional}
\BIBentryALTinterwordspacing
J.~Hewitt, K.~Ethayarajh, P.~Liang, and C.~Manning, ``Conditional probing: measuring usable information beyond a baseline,'' in \emph{Proceedings of the 2021 Conference on Empirical Methods in Natural Language Processing}, M.-F. Moens, X.~Huang, L.~Specia, and S.~W.-t. Yih, Eds.\hskip 1em plus 0.5em minus 0.4em\relax Online and Punta Cana, Dominican Republic: Association for Computational Linguistics, Nov. 2021, pp. 1626--1639. [Online]. Available: \url{https://aclanthology.org/2021.emnlp-main.122}
\BIBentrySTDinterwordspacing

\end{thebibliography}

\end{document}